\documentclass[10pt,twocolumn,letterpaper]{article}

\usepackage{iccv}
\usepackage{times}
\usepackage{epsfig}
\usepackage{graphicx}
\usepackage{amsmath}
\usepackage{amssymb}

\usepackage{subcaption}
\usepackage{booktabs}
\usepackage{multirow}
\usepackage{multicol}
\usepackage{overpic}
\usepackage[final]{animate}

\usepackage{parskip}
\usepackage[font=small,skip=5pt]{caption}
\usepackage{mdwlist}
\usepackage[breaklinks=true,bookmarks=false]{hyperref}

\DeclareRobustCommand\onedotojw{\futurelet\@let@token\onedotojwaux}
\def\onedotojwaux{\ifx\@let@token.\else.\null\fi\xspace}
\providecommand{\eg}{\emph{e.g}\onedotojw} 
\providecommand{\ie}{\emph{i.e}\onedotojw} 
 
\providecommand{\etc}{\emph{etc}\onedotojw} \providecommand{\vs}{\emph{vs}\onedotojw}
\providecommand{\wrt}{w.r.t\onedotojw} 

\providecommand{\etal}{\emph{et al}\onedotojw}
\newcommand{\fig}[1]{Fig.~\ref{fig:#1}}

\newcommand{\image}{I}

\newcommand{\encp}{E}
\newcommand{\enc}{\mathrm{E}}
\newcommand{\dec}{\mathrm{D}}

\newcommand{\gridX}{\mathbf{X}}

\newcommand{\mask}{M}
\newcommand{\sampler}{\mathrm{S}}

\newcommand{\params}{\theta}
\newcommand{\allparams}{\Theta}
\newcommand{\flowfield}{F}
\newcommand{\numfeats}{n}

\newcommand{\ktol}{{k\rightarrow l}}
\renewcommand{\L}{\textsf{\small L}}
\newcommand{\loss}{\mathcal{L}}
\newcommand{\occ}{O}
\newcommand{\seg}{S}

\newcommand{\ith}{\ensuremath{i^\textrm{th}}\xspace}

\iccvfinalcopy

\begin{document}

\title{Transformable Bottleneck Networks}

\newcommand*\samethanks[1][\value{footnote}]{\footnotemark[#1]}

\author{Kyle Olszewski$^1$$^3$\thanks{This work was performed while the author was at Snap Inc.}, Sergey Tulyakov$^2$, Oliver Woodford$^2$, Hao Li$^1$$^3$$^4$, and Linjie Luo$^5$\samethanks\\
  $^1$University of Southern California, $^2$Snap Inc., $^3$USC ICT, $^4$Pinscreen Inc., $^5$ByteDance Inc.}

\maketitle

\begin{abstract}

We propose a novel approach to performing fine-grained 3D manipulation of image content via a convolutional neural network, which we call the Transformable Bottleneck Network (TBN). It applies given spatial transformations directly to a volumetric bottleneck within our encoder-bottleneck-decoder architecture. Multi-view supervision encourages the network to learn to spatially disentangle the feature space within the bottleneck. The resulting spatial structure can be manipulated with arbitrary spatial transformations. We demonstrate the efficacy of TBNs for novel view synthesis, achieving state-of-the-art results on a challenging benchmark. We demonstrate that the bottlenecks produced by networks trained for this task contain meaningful spatial structure that allows us to intuitively perform a variety of image manipulations in 3D, well beyond the rigid transformations seen during training. These manipulations include non-uniform scaling, non-rigid warping, and combining content from different images. Finally, we extract explicit 3D structure from the bottleneck, performing impressive 3D reconstruction from a single input image.~\footnote{Code and data for this project are available on our website: \href{https://github.com/kyleolsz/TB-Networks}{https://github.com/kyleolsz/TB-Networks}}

\end{abstract}
\section{Introduction}
\label{sec:intro}

Inferring and manipulating the 3D structure of an image is a challenging task, but one that enables many exciting applications. By rigidly transforming this structure, one can synthesize novel views of the content. More general transformations can be used to perform tasks such as warping or exaggerating features of an object, or fusing components of different objects. Convolutional Neural Networks (CNNs) have shown impressive results on various 2D image synthesis and manipulation tasks, but specifying such fine-grained and varied 3D manipulations of the image content, while achieving high-quality synthesis results, remains difficult.

Several approaches to providing transformation parameters as an input to, and applying such transformations within, a network have been explored.
A common approach is to pass spatial transformation parameters as an explicit input vector to the network~\cite{tran2017disentangled}, optionally with a decoder trained to perform a specific set of transformations~\cite{eslami2018neural,tatarchenko2015single}. Other approaches include altering the input by augmenting it with auxiliary channels defining the desired spatial transformation~\cite{ma2017pose}, or constructing a renderable representation that is spatially transformed prior to rendering~\cite{liu2018geometry,tulsiani2018layer}.

\begin{figure}
\centering
\includegraphics[width=\linewidth]{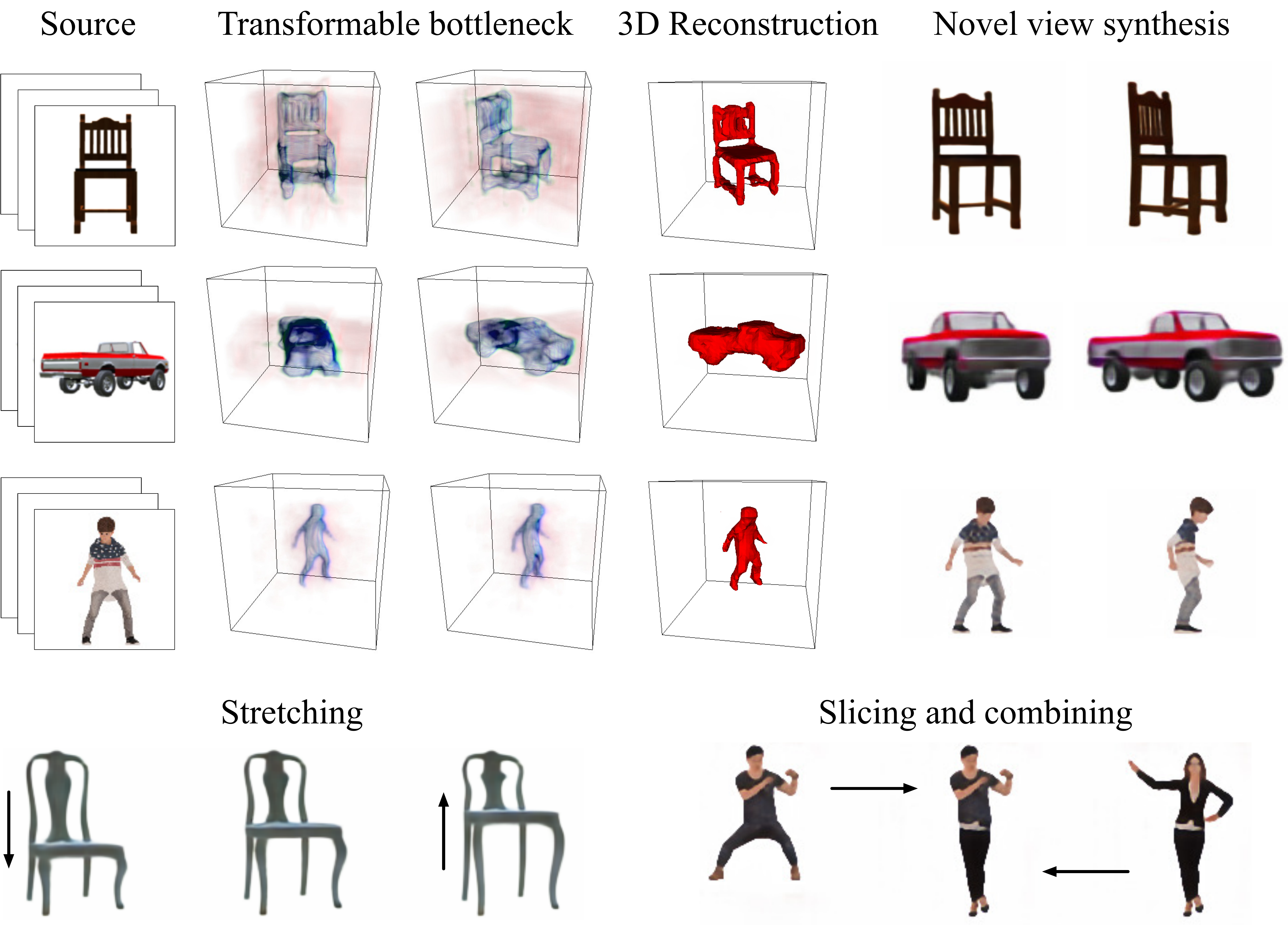}
\caption{\textbf{Applications of TBNs}. A Transformable Bottleneck Network uses one or more images (\emph{column 1}; here, 4 randomly sampled views) to encode volumetric bottlenecks (\emph{columns 2 \& 3}), which are explicitly transformed into and aggregated in an output view coordinate frame.
Transformed bottlenecks are then decoded to synthesize state-of-the-art novel views (\emph{columns 5 \& 6}), as well as reconstruct 3D geometry (\emph{column 4}).
Fine-grained and non-rigid transformations, as well as combinations, can be applied in 3D, allowing creative manipulations (\emph{bottom row}) that were never used during training. Images shown are samples of real results.
}
\label{fig:teaser}
\end{figure}

We propose a novel approach: directly applying the spatial transformations to a volumetric bottleneck within an encoder-bottleneck-decoder network architecture. We call these \emph{Transformable Bottleneck Networks} (TBNs). The network learns that these 3D transformations correspond to transformations between source and target images.

There are several advantages to this approach. Firstly, supervising on multi-view datasets encourages the network to infer spatial structure---it learns to spatially disentangle the feature space within the bottleneck. Consequently, even when training a network using only \emph{rigid} transformations corresponding to viewpoint changes, we can manipulate the network output at test time with \emph{arbitrary} spatial transformations (see Figs.~\ref{fig:teaser} \&~\ref{fig:creative-manipulation}). The operations enabled by these transformations thus include not only rotation and translation, but also effects such as non-uniform 3D scaling and global or local non-rigid warping. Additionally, bottleneck representations of multiple inputs can be transformed into, and combined in, the same coordinate frame, allowing them to be aggregated naturally in feature space. This can resolve ambiguities present in a representation from a single image. While similar to ideas in Spatial Transformer Networks (STN)~\cite{jaderberg2015spatial,lin2017inverse} and a 3D reconstruction method~\cite{rezende2016unsupervised} deriving from it, a key distinction of our approach is that the spatial transformations are input to our network, as opposed to inferred by the network. It is precisely this difference that enables TBNs to make such diverse manipulations.

We highlight the power of this approach by applying it to novel view synthesis (NVS). NVS is a challenging task, requiring non-trivial 3D understanding from one or more images in order to predict corresponding images from new viewpoints. This allows us to demonstrate both the ability of a TBN to naturally spatially disentangle features within a 3D bottleneck volume, and the benefits that this confers.
We compare to leading NVS methods~\cite{tatarchenko2015single,zhou2016flow,sun2018nvs,park2017transformation}, on images from the ShapeNet dataset~\cite{chang2015shapenet}, and attain state-of-the-art results on both $\L_1$ and SSIM metrics (see Table~\ref{tab:nvsresults}, and Figs.~\ref{fig:teaser} \&~\ref{fig:nvs-examples}). We present additional qualitative results on a synthetic human performance dataset. We also train a simple voxel occupancy classifier on image segmentations (\ie without 3D supervision), and use it to demonstrate accurate 3D reconstructions from a single image. Finally, we provide qualitative examples of how this bottleneck structure allows us to perform realistic, varied and creative image manipulation in 3D (Figs.~\ref{fig:teaser} \&~\ref{fig:creative-manipulation}).

In summary, the main contributions of this work are:
\begin{itemize*}
\item A novel, transformable bottleneck framework that allows CNNs to perform spatial transformations for highly controllable image synthesis.
\item A state-of-the-art NVS system using TBNs.
\item A method for extracting high-quality 3D structure from this bottleneck, constructed from a single image.
\item The ability to perform realistic, varied and creative 3D image manipulation.
\end{itemize*}

\section{Related work}
\label{sec:related}

We now review works related to the TBN, in the areas of image and novel view synthesis, and volumetric reconstruction\footnote{Image to depth map~\cite{godard2017monodepth,li2018megadepth}, 3D mesh~\cite{henderson2018learning,jack2018learningffd,wang2018pixel2mesh}, point cloud~\cite{fan2017point} and surfel primitive~\cite{groueix2018atlasnet} approaches also exist, but are outside the scope of our discussion.} and rendering.

\subsection{Image and novel view synthesis}

Many exciting advances in image synthesis and manipulation have emerged recently that enable the application of specific styles or attributes. Early approaches generated natural images using samples from a chosen distribution using a generative adversarial (GAN) training scheme~\cite{goodfellow2014generative,radford2015unsupervised}. Conditional methods then provided the ability to change the style of an input image to another style~\cite{isola2017image,liu2017unsupervised}. Initially such trained networks could only handle one style~\cite{zhu2017unpaired}; more recent works now allow multiple attribute changes using a single network, by learning to disentangle these attributes from the training images~\cite{lample2017fader,tran2017disentangled,zhu2018von}.

Novel view synthesis generates an image from a new, user-specified viewpoint, given one or more images of a scene from known viewpoints. We focus on methods that, like ours, can synthesize novel views from a single input image. This is a highly ill-posed problem, requiring strong 3D understanding and disentanglement of viewpoint and object shape from the input image. Since the seminal work of Hoiem \etal~\cite{hoiem2005popup}, methods have sought to develop more expressive models to address general NVS.
Early CNN solutions regressed output pixel color in the new view~\cite{tatarchenko2015single,yang2015weakly} directly from the input image. Some works disentangle their representations~\cite{tran2017disentangled,yang2015weakly}, separating pose from object~\cite{yang2015weakly} or face identity~\cite{tran2017disentangled}. Zhou \etal~\cite{zhou2016flow} introduced a flow prediction formulation, inferring an output to input pixel mapping instead, to which an explicit occlusion detection and inpainting module~\cite{park2017transformation} and generalization to an arbitrary number of input images~\cite{sun2018nvs} have been added. Eslami \etal~\cite{eslami2018neural} developed a latent representation that can be \emph{aggregated} to combine inputs, showing good results on synthetic geometric scenes.

A drawback of all these approaches is that they condition their networks to perform the transformation, limiting the transformations that can be applied to those that have been learned. Most recently, methods have been proposed to generate explicit representations of geometry and appearance that are transformed and rendered using standard rendering pipelines~\cite{liu2018geometry,tulsiani2018layer}. While these representations can be rendered from arbitrary viewpoints, they are based on planar representations and are therefore not able to capture realistic shape, especially when rendered from side views. Our TBN approach allows us to perform fine-grained and varied, even non-rigid, 3D manipulations in the bottleneck volume, synthesizing them into realistic novel views. Here, the manipulations are applied manually. However, recent work~\cite{wang20193dn} proposes a learned network for deforming objects arbitrarily (parameterized by an input shape), an idea that complements our framework.

\begin{figure*}
	\centering
	\includegraphics[width=0.90\linewidth]{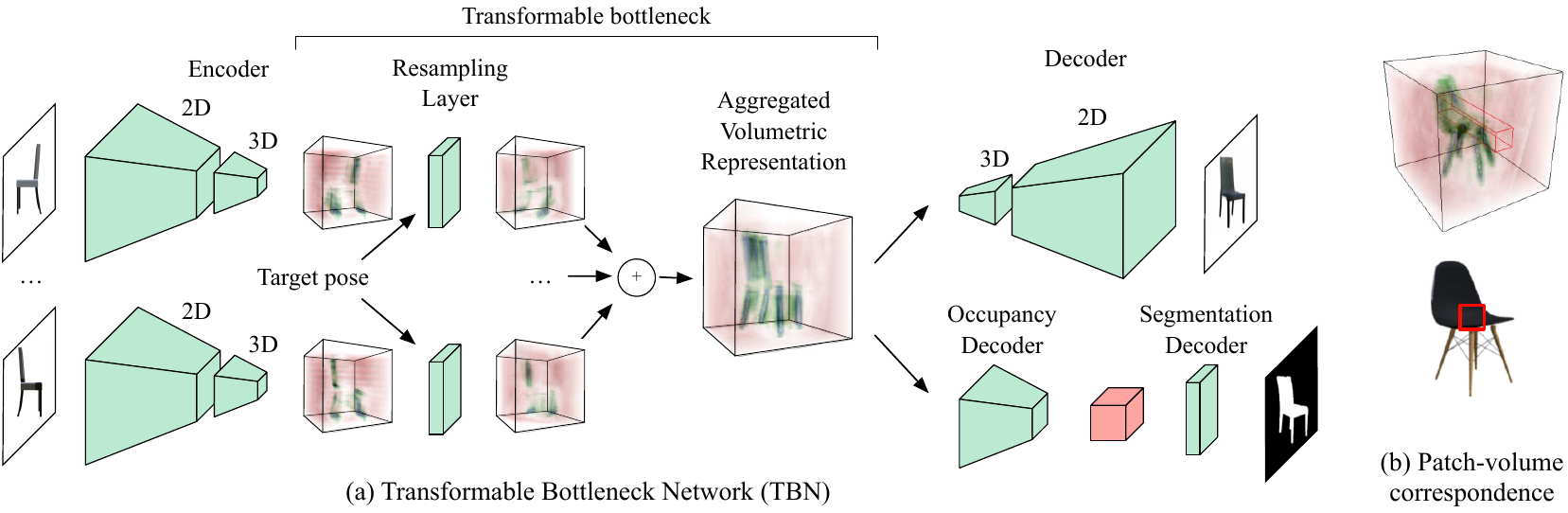}
	\caption{\textbf{A Transformable Bottleneck Network}. (a) Network architecture, consisting of three parts: an encoder (2D convolution layers, reshaping, 3D convolution layers), a resampling layer, and a decoder (a mirror of the encoder architecture). The encoder and decoder are connected purely via the bottleneck; no skip connections are used. The resampling layer transforms an encoded bottleneck to the target view via trilinear interpolation. It is parameterless, \ie transformations are applied explicitly, rather than learned. Multiple inputs can be aggregated by averaging bottlenecks prior to decoding. (b) A visualization of the conceptual correspondence between an image patch and a subvolume of the bottleneck. Bottleneck volume visualizations show the cellwise norm of feature vectors. It is interesting that to note that this norm appears to encode the object shape.}
	\label{fig:framework}
\end{figure*}

\subsection{Volumetric reconstruction and rendering}

Several recent methods reconstruct an explicit occupancy volume from a single image~\cite{choy20163d,girdhar2016predictable,kar2017lsm,rezende2016unsupervised,tulsiani2017ray,wu2016learning,wu2017marrnet,yan2016perspective}, some of which are trained using only supervision from 2D images~\cite{rezende2016unsupervised,tulsiani2017ray,yan2016perspective}. Yan \etal~\cite{yan2016perspective} max-pool occupancy along image rays to produce segmentation masks, and minimize their difference \wrt the ground-truths. Tulsiani \etal~\cite{tulsiani2017ray} enforce photo-consistency between projected color images (given the camera poses) using the correspondences implied by the occupancy volume. In contrast to these approaches that use explicit occupancy volumes and rendering techniques, the implicit approaches proposed by Kar \etal~\cite{kar2017lsm}, and in particular Rezende \etal~\cite{rezende2016unsupervised}, are more relevant to our work---both the volumetric representation and the decoder (rendering) are learned, similar to recent neural rendering work~\cite{nguyen2018rendernet}. The former~\cite{kar2017lsm}, trained on ground truth geometry to estimate geometry from images,\!\!\footnote{The latent representation therefore does not encode appearance.} uses three learned networks\footnote{For 2D image encoding, recurrent fusion and a 3D grid reasoning.} and a hand-designed unprojection step to compute a latent volume. The latter~\cite{rezende2016unsupervised} requires the target transformation to be inferred by the network for NVS, whereas ours requires it to be provided as input, removing any limitations on the transformations that can be applied at test time.

\section{Transformable bottleneck networks}
\label{sec:method}
In this section we formally define our Transformable Bottleneck Network architecture and training method.

\subsection{Architecture}
\noindent
A TBN architecture (\fig{framework}(a)) consists of three blocks: 
\begin{enumerate*}
    \item An encoder network $\enc: \image_k \rightarrow \gridX_k$ with parameters $\params_\encp$, that takes in an image $\image_k$ and, through a series of 2D convolutions, reshaping, and 3D convolutions,\!\!\footnote{See the appendix for the exact architecture.} outputs a bottleneck representation, $\gridX_k$, structured as a volumetric grid of cells, each containing an $\numfeats$-dimensional feature vector.
    \item A \emph{parameterless} bottleneck resampling layer ${\sampler: \gridX_k, \flowfield_\ktol \rightarrow \gridX'_l}$, that takes a bottleneck representation and user-provided transformation parameterization, $\flowfield_\ktol$, as input, and transforms the bottleneck via a trilinear resampling operation.
    \item A decoder network $\dec_\image: \gridX'_l \rightarrow \image'_l$ with parameters $\params_\image$, whose architecture mirrors that of the encoder, that decodes the transformed bottleneck, $\gridX'_l$, into an output image, $\image'_l$.
\end{enumerate*}
Subscripts $k$ and $l$ represent viewpoints.
Neither the encoder nor the decoder are trained to perform a transformation: it is fully encapsulated in the bottleneck resampling layer. As this layer is parameterless, the network cannot \emph{learn} how to apply a particular transformation at all; rather, it is applied explicitly.
A single source image synthesis operation, which is end-to-end trainable, is written as:
\begin{equation}
\label{eqn:framework}
\image'_l = \dec_\image(\sampler(\enc(\image_k, \params_\encp), \flowfield_\ktol), \params_\image).
\end{equation}
When $\flowfield_\ktol$ is the identity transform (\ie $k = l$), this operation defines an auto-encoder network.

\subsubsection{Handling multiple input views}
\label{sec:method:multiple}
Our formulation naturally extends to an arbitrary number of inputs, both for training and testing, without modifications to either encoder or decoder. The encoded and transformed representations of all inputs are simply averaged:
\begin{align}
    \gridX'_{l} = \frac{1}{|K|} \sum_{k\in K} \sampler(\gridX_k, \flowfield_{k \rightarrow l}),
    \label{eq:multiview-decode}
\end{align}
where $K$ is the set of input viewpoints.
The number of inputs tested on can differ from the number trained on, which can differ even within a training batch.
We later show that the model trained with a single input view can effectively aggregate multiple inputs at inference time, and also that a model trained on multiple inputs can perform state-of-the-art inference from a single image.

\subsubsection{Bottleneck layout and resampling}
The network architecture defines the number of cells along each side of the bottleneck volume, but not the spatial position of each cell. Indeed, the framework imposes no constraints on their position, \eg the voxel grid cells do not need to be equally spaced. In this work the grid cells are chosen to be equally spaced,\!\!\footnote{The scale of the spacing is unimportant here, as our NVS experiments only involve camera rotations around the object center.} with the volume centered on the target object and axis aligned with the camera coordinate frame. Perspective effects caused by projection through a pinhole camera, and the camera parameters that affect them (such as focal length), are learned in the encoder and decoder networks, rather than handled explicitly.

Since the bottleneck representation is a volume, it can be resampled via trilinear interpolation, which is fully differentiable~\cite[Eqn. 9]{jaderberg2015spatial}. This allows it to be spatially transformed. The transformation, $\flowfield_\ktol$, is parameterized as a flow field that, for each output grid cell, defines the 3D point in the input volume to sample to generate it. The decoder takes as input a volume of the same dimensions as the encoder produces, therefore the flow field also has these dimensions. Feature channels form separate volumes that are resampled independently, then recombined to form the output volume.

When the view transformation is rigid, as in the case of NVS, the flow field is computed by transforming the cell coordinates of the novel view by the inverse of the relative transformation from the input view.\!\!\footnote{The flow is defined from output voxel to input voxel coordinate.} 
Non-rigid deformations can also be applied, enabling creative shape manipulation, which we demonstrate in Sec.~\ref{sec:results:nonrigid}. Importantly, we do not train on these kinds of transformations.

\subsubsection{Geometry decoder}
Since the TBN spatially disentangles shape and appearance within the volumetric bottleneck, it should also be able to reconstruct an object in 3D from the bottleneck representation. Indeed, prior work~\cite{rezende2016unsupervised,tulsiani2017ray} shows that training a 3D reconstruction using the NVS task alone, \ie without 3D supervision, is possible. We extract shape in the form of a scalar occupancy volume, $\occ$, with one value per bottleneck cell, using a separate, shallow network, occupancy decoder, $\dec_\occ : \gridX \rightarrow \occ$. 
To avoid using any 3D supervision to train this decoder, we then apply another decoding layer, $\dec_\seg : \occ \rightarrow \seg$, that applies a 1D convolution along the $z$-axis (the optical axis), followed by a sigmoid, to generate a scalar segmentation image $\seg$, thus:
\begin{align}
    \label{eq:occupancy_decoder}
    \seg = \dec_\seg(\occ, \params_\seg), ~~~~ \occ = \dec_\occ(\gridX, \params_\occ),
\end{align}
where $\params_\occ$ and $\params_\seg$ are the parameters of the occupancy and segmentation decoders respectively.

\subsection{Training}
\label{sec:learning}
We train the TBN using the NVS task as follows. 

\subsubsection{Appearance supervision}
NVS requires a minimum of two images of a given object from different, known viewpoints.\!\!\footnote{Viewpoints are defined by camera rotation and translation, \wrt some arbitrary reference frame; world coordinates are not required.}
Given $\{ \image_k$, $\image_l \}$ and $\flowfield_\ktol$, we can compute a reconstruction, $\image'_l$, of $\image_l$ using equation (\ref{eqn:framework}). Using this, we define several losses in image space with which to train our network parameters. The first two are a pixel-wise $\L_1$ reconstruction loss and an $\L_2$ loss in the feature space of the VGG-19 network, often referred to as the perceptual loss:
\begin{align}
    \label{eq:l1-loss}
    \loss_{\mathrm{R}} (\params_\encp, \params_\image) &= || \image_{k \rightarrow l} - \image_l ||_1,\\
    \label{eq:perception-loss}
    \loss_{\mathrm{P}} (\params_\encp, \params_\image) &= \sum_i || \mathrm{V}_i(\image_{k \rightarrow l}) - \mathrm{V}_i(\image_l)||_2^2, 
\end{align}
where $\mathrm{V}_i$ is the output of the \ith layer of the VGG-19 network. To enforce structural similarity of the outputs we also adopt the structural similarity loss~\cite{snell2017msssim,wang2004ssim}, denoted as $\loss_\mathrm{S}$. Finally, we employ the adversarial loss of Tulyakov \etal~\cite{tulyakov2018mocogan}, $\loss_\mathrm{A}$, to increase the sharpness of the output image.

\subsubsection{Segmentation supervision}
Appearance supervision is sufficient for NVS tasks, but to compute a 3D reconstruction we also require segmentation supervision,\!\!\footnote{3D supervision could be used, but requires ground truth 3D data.} in order to learn $\params_\occ$ and $\params_\seg$. We therefore assume that for each image $\image_i$ we also have a binary mask $\mask_i$, with ones on the foreground object pixels and zeros elsewhere.\!\!\footnote{Segmentation supervision is not a hard constraint, therefore segmentations from state-of-the-art methods (\eg Mask R-CNN~\cite{he2017maskrcnn}) may suffice. However, we use ground truth masks in this work.} Segmentation losses are computed in \emph{all} input and output views, using the aggregated bottleneck in the multi-input case, as follows:
\begin{align}
    \loss_\mathrm{M}(\params_\encp,\params_\occ,\params_\seg) = \sum_{k\in K} & H (\dec_\seg(\sampler( \occ_l , \flowfield_{l \rightarrow k}), \params_\seg), \mask_k),\nonumber\\
    +~&H(\dec_\seg(\occ_l, \params_\seg), \mask_l),
    \label{eq:multiview-decode}
\end{align}
where $\occ_l = \dec_\occ(\gridX'_l, \params_\occ)$ and $H$ is the binary cross entropy cost, summed over all pixels. Summing over all views achieves a kind of space carving. Correctly reconstructing unoccupied cells within the visual hull is difficult to learn as no 3D supervision is used, but appearance supervision helps address this.

\subsubsection{Optimization}
The total training loss, with hyper-parameters $\lambda_i$ to control the contribution of each component, is
\begin{align}
    \loss_{\mathrm{T}}(\allparams) = \loss_{\mathrm{R}} + \lambda_1 \loss_{\mathrm{P}}  + \lambda_2 \loss_\mathrm{S} + \lambda_3 \loss_\mathrm{A} + \lambda_4 \loss_\mathrm{M},
    \label{eq:total-loss}
\end{align}
This loss is fully differentiable, and the network can be trained end-to-end by minimizing the loss \wrt the network parameters $\allparams=\{\params_\encp,\params_\image,\params_\occ,\params_\seg\}$ using gradient descent. 

\begin{figure*}
  
  \begin{subfigure}[t]{0.67\textwidth}
  \begin{flushleft}
      \includegraphics[height=6.2cm]{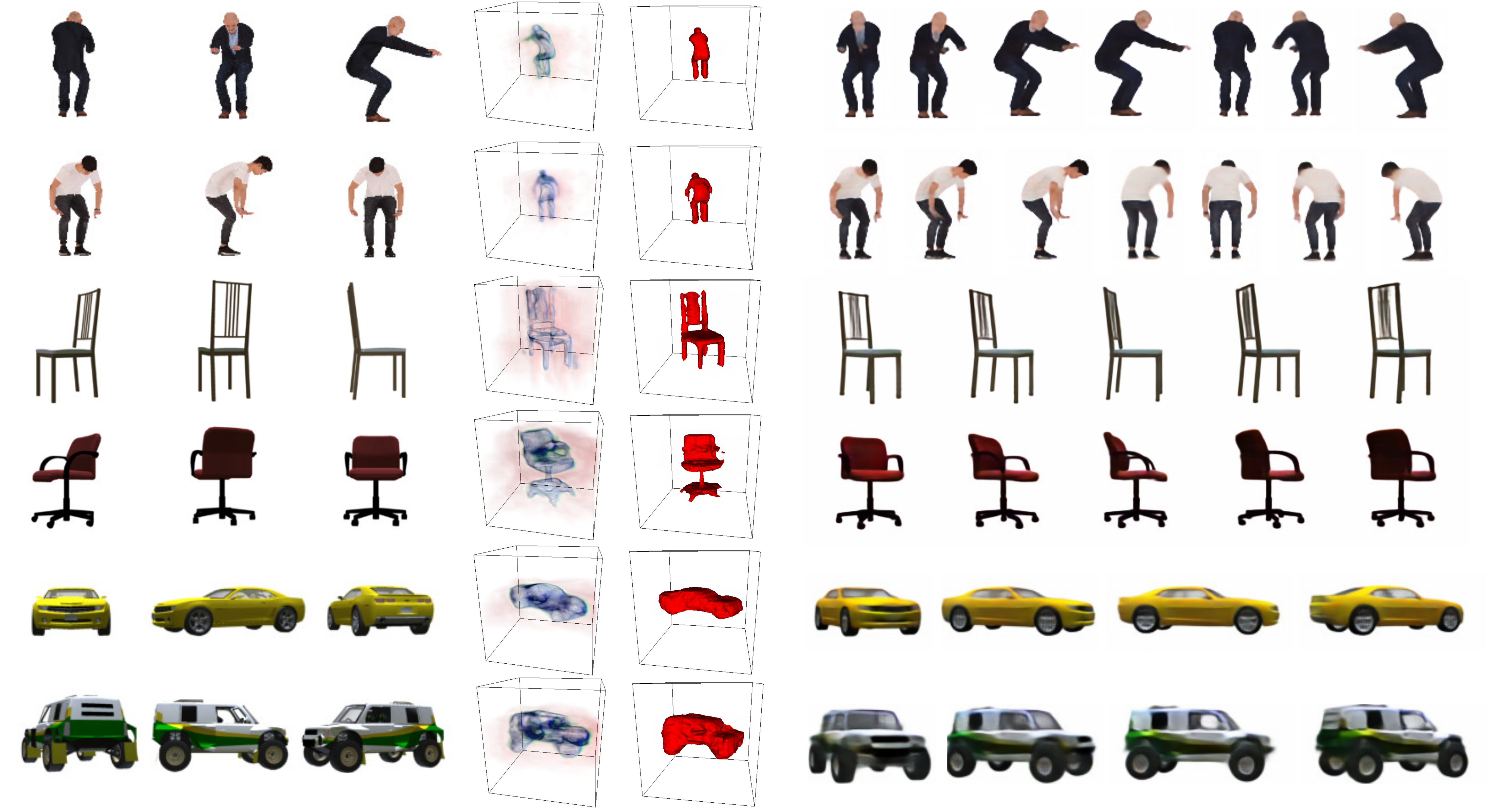}
      \subcaption{Results on novel view synthesis}
      \label{fig:nvs-examples}
  \end{flushleft}
  \end{subfigure}
  \begin{subfigure}[t]{0.33\textwidth}
    \begin{flushright}
      \includegraphics[height=6.2cm]{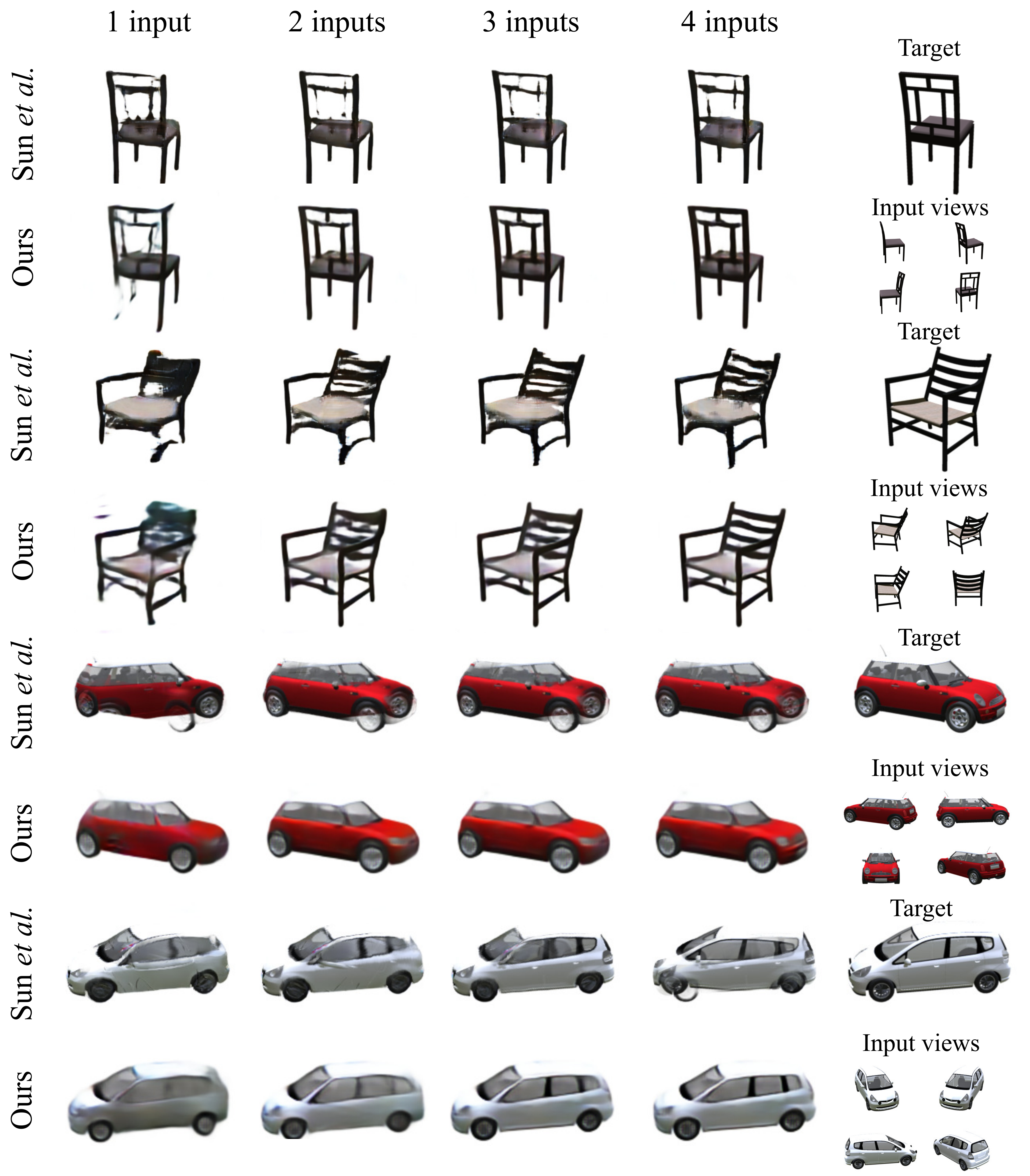}
      \subcaption{Comparisons with Sun \etal~\cite{sun2018nvs}}
      \label{fig:qual_compare_nvs_00}
    \end{flushright}
  \end{subfigure}
  \caption{\textbf{Qualitative results and comparisons.} (\subref{fig:nvs-examples}) Randomly selected NVS samples generated using our method. \emph{Left}: input images (3 of the 4 used). \emph{Middle}: transformable bottleneck and 3D reconstruction. \emph{Right}: synthesized output views. (\subref{fig:qual_compare_nvs_00}) Samples of synthesized novel views using the method of Sun \etal~\cite{sun2018nvs} and ours. Their method fails to capture overall structure for chairs, and generates unnatural artifacts on cars, especially around the wheels. Where $<4$ input views are used, they are selected in clockwise order, starting top left.}
  \label{fig:qualitative-figure}
\end{figure*}

\section{Experiments}
\label{sec:results}

We train and evaluate our framework on a variety of tasks. We provide quantitative evaluations for our results for novel view synthesis using both single and multi-view input, and compare our results to state-of-the-art methods on an established benchmark. We also perform 3D object reconstruction from a single image and quantitatively compare our results to recent work~\cite{tulsiani2017ray}. Finally, we provide qualitative examples of our approach applying creative manipulations via non-rigid deformations.

\subsection{A note on implementation}
Our models are implemented and trained using the  PyTorch framework \cite{paszke2017automatic}, for automatic differentiation and parallelized computation for training and inference. We extended this framework to include a layer to perform parallelizable trilinear resampling of a tensor, in order to efficiently perform our spatial transformations. We plan to release the source code for our framework to the research community upon publication.

Each network was trained on 4 NVIDIA P100s, with each batch distributed across the GPUs. As we found that batch size had no discernible effect on the final result, we selected it to maximize GPU utilization. We trained each model until convergence on the test image set, which took approximately 8 days. For more details on the network architecture, training process and datasets used in our evaluations and results, please consult the appendix.

\subsection{Novel view synthesis}
\label{sec:results:nvs}

\textbf{Setup.} We use renderings of objects obtained from the ShapeNet~\cite{chang2015shapenet} dataset, which provides textured CAD models from a variety of object categories. We measure the capability of our approach to synthesize new views of objects under large transformations, for which ground-truth results are available. We train and evaluate our approach using the cars and chairs categories, to demonstrate its performance on objects with different structural properties. Each model is rendered as $256 \times 256$ RGB images at 18 azimuth angles sampled at 20-degree intervals and 3 elevations (0, 10 and 20 degrees), for a total of 54 views per model. We use standard training and test data splits~\cite{park2017transformation,sun2018nvs,zhou2016flow}, and train a separate network for each object category (also standard), using 4 input images to synthesize the target view. The network architecture and training method were fixed across categories.

\begin{table}
  \centering
  \resizebox{0.40\textwidth}{!}{
  \def\sym#1{\ifmmode^{#1}\else\(^{#1}\)\fi}
  \begin{tabular}{c*{6}{c}}
    \toprule
    & Methods & \multicolumn{2}{c}{Car}  & \multicolumn{2}{c}{Chair} \\
    \cmidrule(lr){3-4}\cmidrule(lr){5-6}
    & & \multicolumn{1}{c}{$\L_1$} & \multicolumn{1}{c}{SSIM} & 
      \multicolumn{1}{c}{$\L_{1}$} & \multicolumn{1}{c}{SSIM} \\
    \midrule
    \multirow{5}{*}{\rotatebox[origin=c]{90}{1 view}} & Tatarchenko \etal 2015~\cite{tatarchenko2015single} & .139 & .875 & .223 & .882       \\  
    & Zhou \etal 2016~\cite{zhou2016flow} & .148 & .877 & .229 & .871       \\    
    & Park \etal 2017~\cite{park2017transformation} & .119 & .913 & .202 & .889       \\
    & Sun \etal  2018~\cite{sun2018nvs}& .098 & .923 & .181 & \textbf{.895}       \\
    & Ours &  \textbf{.091}    &  \textbf{.927}    &  \textbf{.178}     &  \textbf{.895}       \\
    \midrule
    \addlinespace
    \multirow{4}{*}{\rotatebox[origin=c]{90}{2 views}} & Tatarchenko \etal 2015~\cite{tatarchenko2015single}    & .124 & .883 & .209  & .890       \\  
    & Zhou \etal 2016~\cite{zhou2016flow}  & .107 &  .901 & .207 &  .881       \\    
    & Sun \etal 2018~\cite{sun2018nvs}  & .078 & .935 & .141 & .911       \\
    & Ours &  \textbf{.072}    &  \textbf{.939}     &  \textbf{.136}    &  \textbf{.928} \\
    \midrule
    \addlinespace
    \multirow{4}{*}{\rotatebox[origin=c]{90}{3 views}} & Tatarchenko \etal 2015~\cite{tatarchenko2015single}    &  .116 & .887 & .197 & .898       \\  
    & Zhou \etal 2016~\cite{zhou2016flow}    &  .089 & .915 & .188 & .887       \\    
    & Sun \etal 2018~\cite{sun2018nvs}   & .068 & .941 & .122 & .919       \\
    & Ours &  \textbf{.063}    &  \textbf{.943}     &  \textbf{.116}    &  \textbf{.936} \\
    \midrule
    \addlinespace
    \multirow{4}{*}{\rotatebox[origin=c]{90}{4 views}} & Tatarchenko \etal 2015~\cite{tatarchenko2015single}    &  .112 & .890 & .192 & .900      \\  
    & Zhou \etal 2016~\cite{zhou2016flow}    &  .081 & .924 & .165 & .891       \\    
    & Sun \etal 2018~\cite{sun2018nvs}   &  .062    &  \textbf{.946}    &  .111     &  .925       \\
    & Ours &  \textbf{.059}    &  \textbf{.946}     &  \textbf{.107}    &  \textbf{.939} \\
    \bottomrule
  \end{tabular}}
\caption{\textbf{Quantitative results on novel view synthesis}. We report the $\L_{1}$ loss (lower is better) and the structural similarity (SSIM) index (higher is better) for our method and several baseline methods, for 1 to 4 input views, on both car and chair ShapeNet categories.}
\label{tab:nvsresults}
\end{table}
 
As described in Section~\ref{sec:method:multiple}, our framework can use a variable number of input images. Though trained with 4 input images, we demonstrate that our networks can infer high-quality target images using fewer input images at test time. Using the experimental protocol of Sun~\etal~2018~\cite{sun2018nvs}, which uses up to 4 input images to infer a target image, we report quantitative results for our approach and others that can use multiple input images~\cite{sun2018nvs,tatarchenko2015single,zhou2016flow}, as well as for an approach accepting single inputs~\cite{park2017transformation}.

To further demonstrate the applicability of our method to non-rigid objects with higher pose diversity and lower appearance diversity, we also train and qualitatively evaluate a network using a multi-view human action dataset~\cite{Renderpeople:2018:RP}. This dataset uses a limited number (186) of textured CAD models representing human subjects.
However, the subjects are rigged to perform animation sequences representing a variety of common activities (running, waving, jumping, \etc), resulting in a much larger number of renderings. Note that the training process is identical to that used for rigid objects---input images for a given scene see the subject in a \emph{fixed} pose.
Thus, the capability to perform non-rigid transformations, as seen in Sec.~\ref{sec:results:nonrigid}, is still implicitly learned by the network.

\textbf{Results.} Table~\ref{tab:nvsresults} reports quantitative results across recent methods, for 1 to 4 input views, on car and chair categories, for both the $\L_{1}$ cost and structural similarity (SSIM) scores~\cite{wang2004ssim}. Though our networks are trained using exactly 4 input views, we obtain state-of-the-art results across \emph{all} metrics, categories and number of input views, even in the challenging case of single-view input.

These results indicate that the TBN excels at NVS, and outperforms alternatives using both pixelwise and perceptual metrics. We further note that our method performs significantly better than others in cases involving large transformations of the input images and challenging viewpoints (see Fig.~\ref{fig:qual_compare_nvs_00}). This demonstrates that our approach to combining information from these viewpoints is an effective strategy for synthesizing novel viewpoints, in addition to having other interesting applications (see below).

Fig.~\ref{fig:nvs-examples} shows qualitative examples on 3 datasets: the ShapeNet cars and chairs used for our quantitative evaluations, and the aforementioned human activity dataset. Fig.~\ref{fig:qual_compare_nvs_00} qualitatively compares our results with those of Sun \etal~\cite{sun2018nvs} on several challenging examples requiring large viewpoint transformations from the chair and car datasets. Their method has difficulty inferring the proper correspondence between the source and target images for both object categories, particularly the more complex and variable structure of the chairs. Thus, many details are missing or incorrectly transformed. For cars, errors in the correspondence between local regions of source and target images cause artifacts, such as the wheel on the front of the car in row 5. In contrast, our method recovers the overall structure of both chairs and cars well, improving finer details as additional input views are added. 
We note that their results are in some cases sharper, as they use flow prediction to directly sample input pixels to construct the output, whereas our output images are rendered entirely from the bottleneck representation, as is required for general 3D manipulation.

\subsection{Appearance synthesis for 3D reconstruction}
\label{sec:results:appearance}
As reported above, our method performs well on NVS with a single view, and progressively improves as more input views are used. We now show that this trend extends to 3D reconstruction. However, given that more views aid reconstruction, and that our network can generate more views, an interesting question is whether the generative power of our network can be used to aid the 3D reconstruction task. We ran experiments to find out.

\textbf{Setup.} To evaluate our method, we use the 3D reconstruction evaluation framework from the Differentiable Ray Consistency (DRC) work of Tulsiani~\etal~\cite{tulsiani2017ray}, which infers a 3D occupancy volume from a single RGB image. We trained our network on their dataset: multi-view images of ShapeNet objects, rendered under varying lighting conditions from 10 viewpoints, randomly sampled from uniform azimuth and elevation distributions with ranges $[0,360)$ and $[-20,30]$, respectively.  As our method is trained using a set of multi-view images and corresponding segmentation masks, we compare our method to their publicly available model trained on masked, color images, using 5 random views of each object. In contrast, for this task our model was trained using only 2 random views (one input, one output) of each object.

Using the DRC~\cite{tulsiani2017ray} experimental protocol, we report the mean intersection-over-union (IoU) of the volumes from our occupancy decoder, computed on the evaluation image set, compared to the ground-truth occupancies obtained by voxelizing the 3D meshes used to render these images. Like DRC, we report the IoU attained using the optimal discretization threshold for each object category.

\begin{figure}
    \centering
    \includegraphics[width=0.89\linewidth]{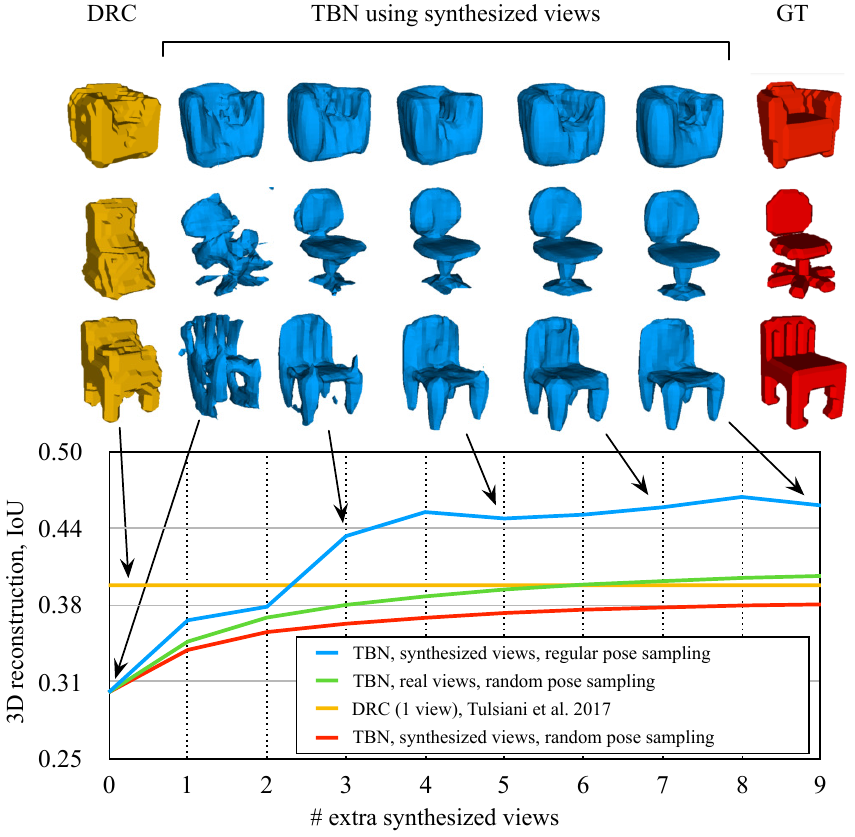}
    \caption{\textbf{3D reconstruction results}. Quantitative (IoU, following the evaluation framework of Tulsiani \etal~\cite{tulsiani2017ray}) and qualitative results of our method performing 3D reconstruction on the chairs dataset, from a single input image, supplemented by additional views synthesized by our network. 0 synthesized views indicates that only the original input image is used, while 1 to 9 indicate that we synthesize these additional views and combine the bottlenecks generated from these viewpoints with those obtained from the original input view. Results from Tulsiani~\etal~\cite{tulsiani2017ray}, who use only one image during inference, are also shown.}
    \label{fig:graph_recon}
\end{figure}

\textbf{Results.} Figure~\ref{fig:graph_recon} shows the results of this evaluation. We report IoU numbers obtained using one real input image, with 0 to 9 additional synthesized views, sampled either randomly (red line) or regularly (at $0^{\circ}$ elevation, blue line). For comparison, we show results using additional real images of the target object (green line), randomly sampled from the evaluation set (regularly sampled images were not available), as well as the results using DRC~\cite{tulsiani2017ray} with a single input image (yellow line).
The figure also contains qualitative comparisons of results\footnote{We render the voxel grids as meshes using an isosurface method.} using our best method (regularly sampled synthetic images) with varying numbers of synthetic images (middle columns), compared to DRC~\cite{tulsiani2017ray} (left) and the ground truth (right). Our method produces good results even with concavities (Fig.~\ref{fig:graph_recon}, row 1), that could not be obtained solely from the object's silhouette, demonstrating that NVS supervision is an able substitute for geometry supervision when inferring the geometric structure of such objects.

Using synthesized views from random poses clearly improves the reconstruction quality as more views are incorporated into our representation, though does not match the quality attained when using the same number of real images instead. Using synthetic views sampled at \emph{regular} intervals around the object's central axis produces significantly better results, achieving superior single view 3D reconstruction to all other methods when using as few as 3 synthetic views.
This dramatic improvement from randomly to regularly sampled synthetic views can be explained by the fact that information from each of the regularly sampled views is much more complementary than for the random views, that could leave parts of the object ``unseen'' (or unhallucinated). That synthetic views should improve the results at all is a more nuanced argument.

One might imagine that recycling hallucinated views into the encoder would simply reinforce the existing reconstruction. However, we argue the following:
the encoder learns to extract the features that allow an image to be transformed, and the decoder learns to process the transformed features so as to produce a plausible image under this transformation. Therefore, consider a chair viewed from only one angle: the encoder could say where in space it believes the \emph{visible} parts be, allowing it to be transformed, then the decoder could see this partial reconstruction in the bottleneck, and knowing what chairs look like, hallucinate the \emph{unseen} parts. By recycling the synthesized image back through the encoder, it could then see new parts of the chair, and generate structure for them also. In essence, it comes down to where unseen structure is hallucinated within the network. Since the bandwidths of our encoder and image decoder are identical, there is no reason for it be in any particular part. However, because the gradients in the decoder layers have been passed through fewer other layers, they may receive a stronger signal for hallucination from the output view, hence learn it first.

One might expect the occupancy decoder to learn to hallucinate structure as well as the image decoder, but our results indicate that it doesn't (see our qualitative reconstructions with no synthetic views, in Fig.~\ref{fig:graph_recon}). We intuit that this is because it has much less information (binary \vs color images) to train on, and concomitantly a significantly smaller bandwidth. This further validates our hypothesis that appearance supervision improves 3D reconstruction within the visual hull, in the absence of 3D supervision.

\begin{figure}
    \centering
    \includegraphics[width=1.00\linewidth]{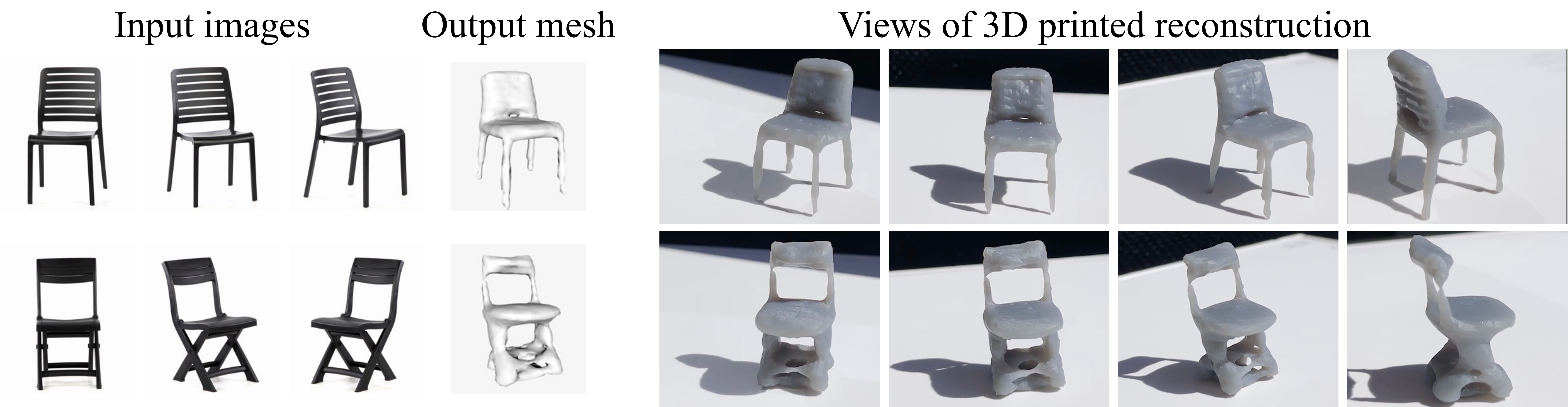}
    \caption{\textbf{Examples of 3D printed objects} created using our approach to 3D reconstruction.}
    \label{fig:3d_examples}
\end{figure}

\textbf{Physical recreations of real objects.} An exciting possibility of image-based reconstruction is being able to recreate old objects from photographs. We took 3 photos each of 2 \emph{real} chairs, computed TBNs from these images and aggregated them using estimated relative poses. We computed occupancy volumes from these, extracted meshes using an isosurface method, and 3D printed these meshes. Figure~\ref{fig:3d_examples} shows the input images, reconstructed meshes and 3D printed objects. Despite the low resolution of the occupancy volume ($40^3$ voxels), these physical recreations are coherent and depict the salient details of each chair.

\subsection{Non-rigid transformations}
\label{sec:results:nonrigid}

\textbf{Spatial disentanglement.} Due to the convolutional nature of our network, a subvolume of the 3D bottleneck broadly corresponds to a patch of the input (if encoding) or output (if decoding) image, as visualized in Fig.~\ref{fig:framework}(b).
Any of the features in the subvolume, or a combination of them, can account for the appearance of the image patch; there is no guarantee that the features used will come from the voxels corresponding to the location in 3D space of the surface seen in the patch. In our framework, however, 2D supervision from multiple directions (both input and output views) places multiple subvolume constraints on where information can be stored. Storing information in the cells corresponding to the location in 3D space of the visible surface is the most efficient layout of information that meets all of those constraints, thus the one which achieves the lowest loss given the available network bandwidth. The effect is therefore achieved implicitly, rather than explicitly. 

\begin{figure}
    \centering

	\makebox[0mm][s]{
			\includegraphics[width=0.44\textwidth]{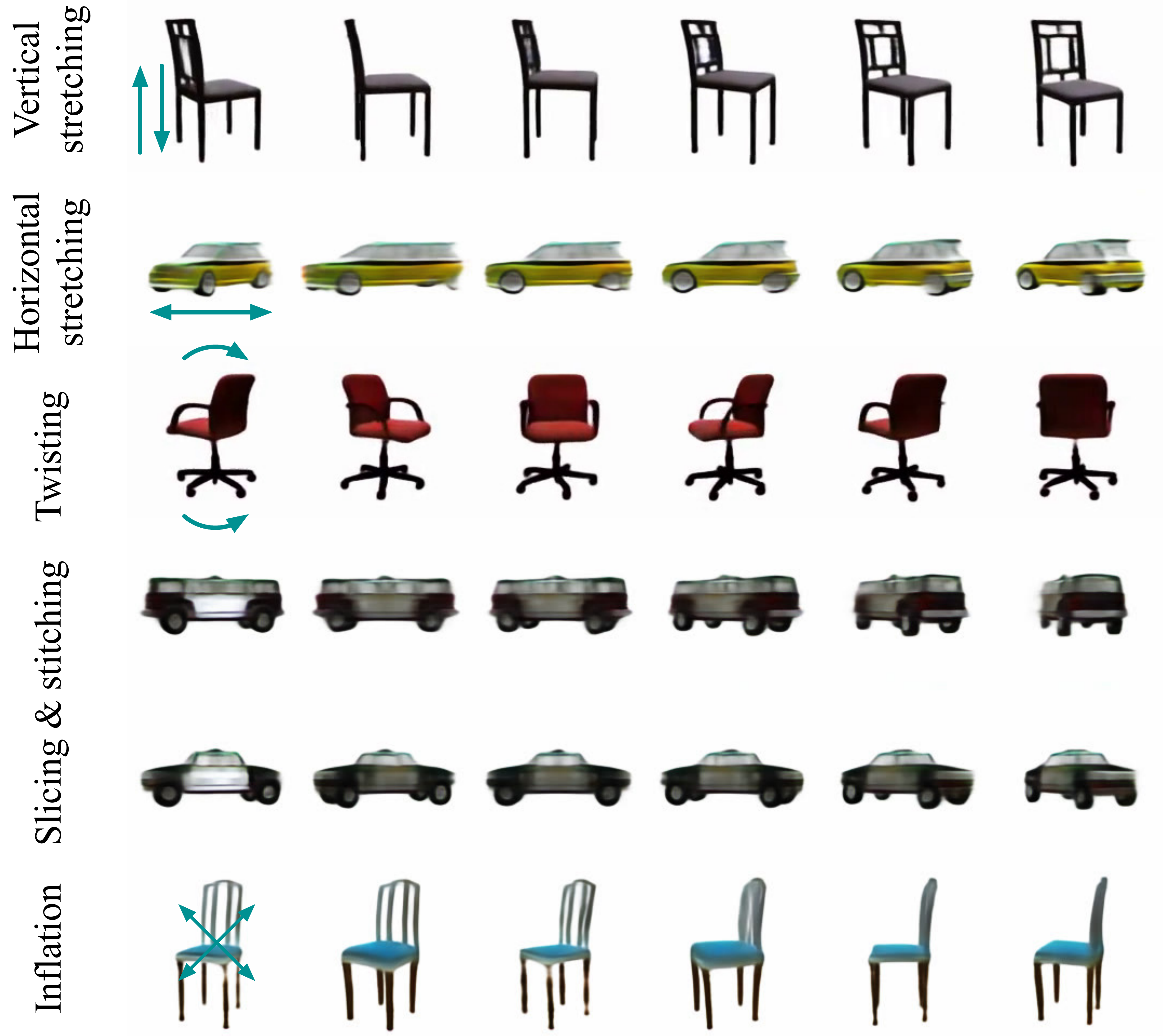}
		}
	\animategraphics[width=0.44\textwidth,autoplay,loop,poster=none]{5}{figures/animated/manipulations/frame_}{0001}{0039}
	\caption{\textbf{Creative, non-rigid manipulations} (\emph{view as videos in Acrobat Reader}). Selected examples of non-rigid 3D manipulations applied to transformable bottlenecks, for creative image and video\protect\footnotemark\,synthesis. Manipulations include: vertical and horizontal stretching, twisting, slicing \& stitching, non-linear inflation. Please see the supplementary video for further demonstrations.
	}
	\label{fig:creative-manipulation}
\end{figure}

\textbf{Creative manipulation.} Based on this effect of spatial disentanglement, \emph{arbitrary} non-rigid volumetric deformations can be applied on the transformable bottleneck, resulting in a similar transformation of shape of the rendered object. We demonstrate this qualitatively with a variety creative tasks, shown in Figure~\ref{fig:creative-manipulation}, that are performed by manipulating and combining the volumetric bottlenecks extracted from input images. Objects can be stretched in different dimensions (\emph{first and second rows}). By rotating the upper and lower portion of the volume in opposite directions (\emph{third row}), we can transform different regions of the target into a new shape that does not correspond to a single rigid transformation. Non-uniform and/or local scaling can be applied to inflate or shrink (\emph{bottom row}) objects. Parts of a bottleneck can even be replaced with another part from the same, or a different bottleneck, creating hybrid objects (\emph{fourth and fifth rows}). Many other such manipulations are possible, far beyond the scope of the rigid transformations trained on.

\footnotetext{While some such manipulations could \emph{seem} simple to achieve in 2D, an edited 3D object can also be rendered consistently from any azimuth (see videos here and in the supplementary video), from a \emph{single} manipulated bottleneck.}

\begin{figure}
    \centering
    \includegraphics[width=1.00\linewidth]{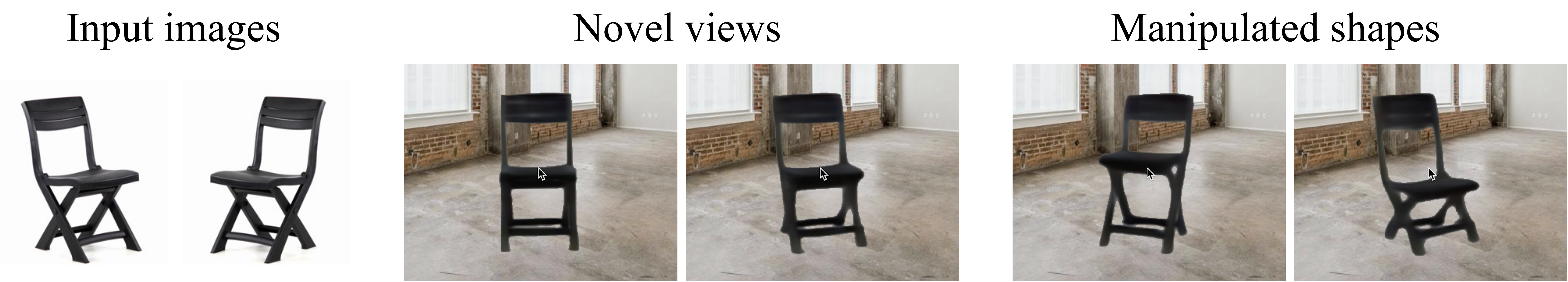}
    \caption{\textbf{Interactive manipulation}. We use our approach to rotate and deform objects before compositing them into real images.}
    \label{fig:manip_examples}
\end{figure}

\vfill\eject
\textbf{Interactive creative manipulation.} We implemented a tool to demonstrate a useful real-world application of the TBN: interactive manipulation and compositing. The user has one or more\footnote{Multiple images require true or estimated relative poses.} photos of an object (whose class has been trained on) they wish to manipulate and place in a photo of a real world scene. The images are loaded into our application, from which a single aggregated bottleneck is computed. An interactive interface then allows the user to rotate, translate, scale and stretch the object, transforming and rendering the bottleneck in realtime and overlaying the object in the target image, as they apply the transformations.

Figure~\ref{fig:manip_examples} contains example inputs and outputs of this process, for an interior design visualization use case. Two photos of a real chair were provided (with estimated relative pose). Rotations and stretches were then applied interactively, to get a feel for how the chair would look with different orientations and styles. Despite the challenging nature of this example (real photos of a chair with complex structure, and real-world lighting conditions such as specular highlights), we achieve highly plausible results.

\section{Conclusion}
\label{sec:conclusion}
This work has presented a novel approach to applying spatial transformations in CNNs: applying them directly to a volumetric bottleneck, within an encoder-bottleneck-decoder network that we call the Transformable Bottleneck Network.
Our results indicate that TBNs are a powerful and versatile method for learning and representing the 3D structure within an image. Using this representation, one can intuitively perform meaningful spatial transformations to the extracted bottleneck, enabling a variety of tasks.

We demonstrate state-of-the-art results on NVS of objects, producing high quality reconstructions by simply applying a rigid transformation to the bottleneck corresponding to the desired view.
We also demonstrate that the 3D structure learned by the network when trained on the NVS task can be straightfowardly extracted from the bottleneck, even without 3D supervision, and furthermore, that the powerful generative capabilities of the complete encoder-decoder network can be used to substantially improve the quality of the 3D reconstructions by re-encoding regularly spaced, synthetic novel views.
Finally, and perhaps most intriguingly, we demonstrate that a network trained on purely rigid transformations can be used to apply arbitrary, non-rigid, 3D spatial transformations to content in images.

{\small
\bibliographystyle{ieee_fullname}
\bibliography{references}
}

\clearpage

\renewcommand{\thesection}{A\arabic{section}}
\setcounter{section}{0}

\section{Architecture}

The overall architecture of our novel view synthesis network is depicted in Table~\ref{tab:arch}. In this table and the corresponding diagrams, \textit{conv} indicates a standard convolutional layer of the specified filter size and stride\footnote{In the text, table and following diagrams, \textit{conv} blocks use a filter size of $3\times3$ and stride $1$, except when otherwise noted.}. In our model, these layers are followed by a batch normalization operation. \textit{upconv} indicates a nearest-neighbor upsampling operation that increases the output width and height by a factor of $2$, followed by a convolution with filter size $3\times3$ and stride $1$, which produces an output of the same size\footnote{Padding is used as necessary to maintain the output dimensions specified at each layer.}, and a batch normalization operation. The \textit{reshape} operation is used before and after the 3D block to produce outputs that match the specified dimensions. \textit{output} is a layer in which a $3\times3$ convolution with stride 1 is applied, followed by a sigmoid operation that produces output in the range of 0 to 1 in each channel. The final output is an RGB image with an additional channel for the segmentation mask. 

\begin{figure}
    \centering
    \includegraphics[width=1.00\linewidth]{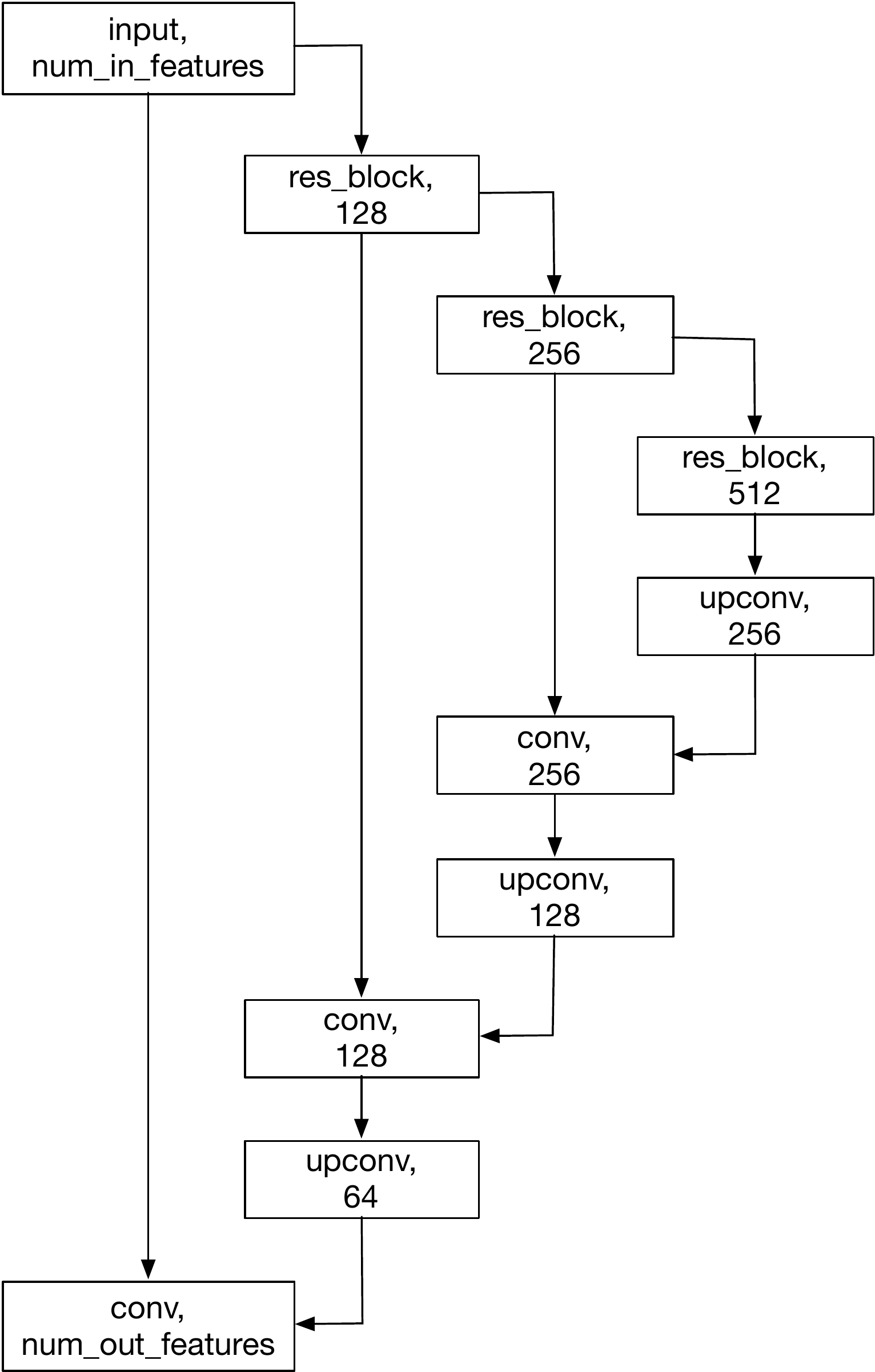}
    \caption{The architecture of the \textit{unet\_block} layers of our model. The number at the bottom of each block indicates the number of feature maps produced by the operation.}
    \label{fig:arch_unet_block}
\end{figure}

\begin{figure}
    \centering
    \includegraphics[width=0.75\linewidth]{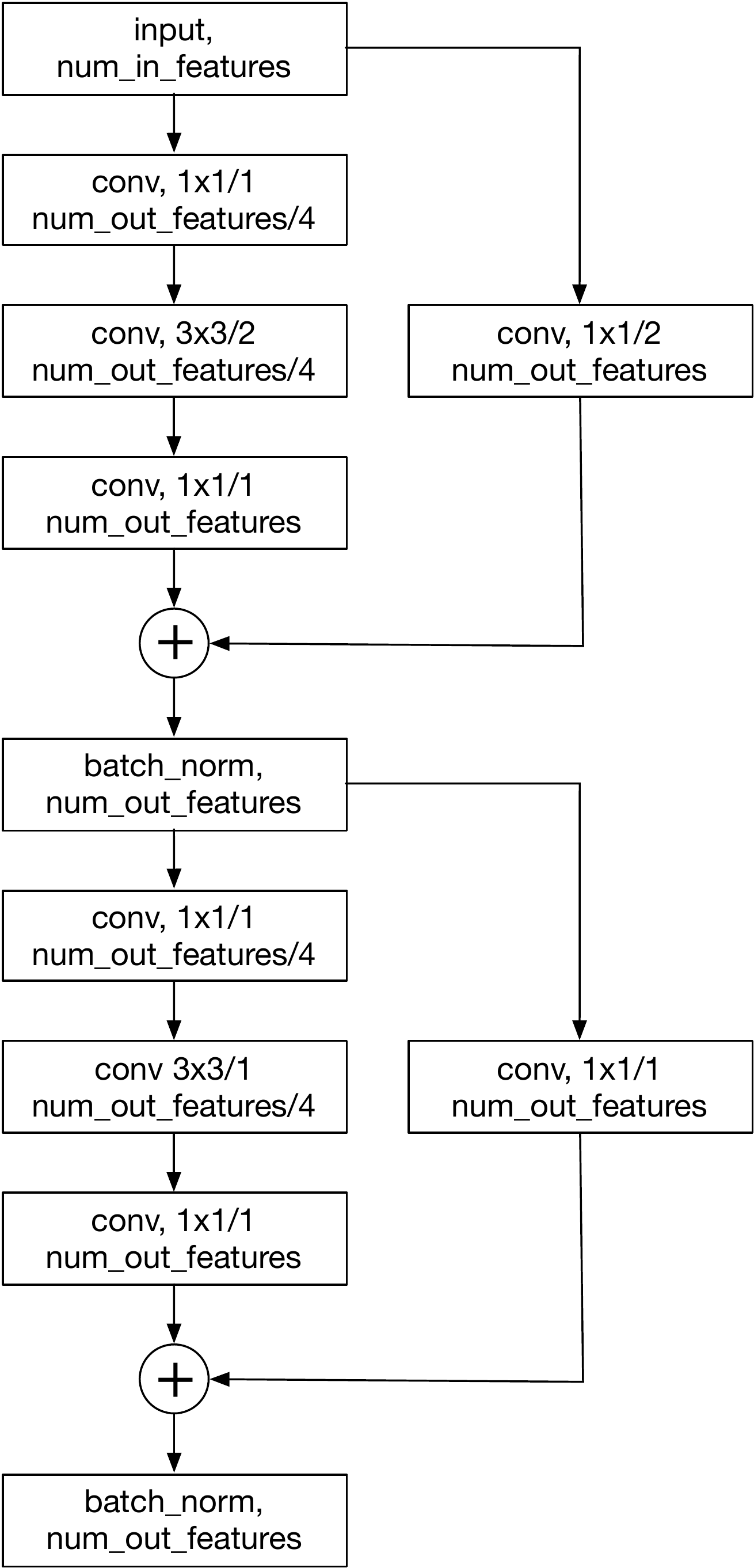}
    \caption{The architecture of the \textit{res\_block} layers used in our architecture, as seen in Fig.~\ref{fig:arch_unet_block}. The top row of the \textit{conv} blocks indicates the filter size and stride, while the number at the bottom of each block indicates the number of feature maps produced by the operation.}
    \label{fig:arch_res_block}
\end{figure}

The architecture of the \textit{unet\_block} segments is depicted in Fig.~\ref{fig:arch_unet_block}. This component uses a standard U-Net architecture~\cite{DBLP:journals/corr/RonnebergerFB15} with skip connections connecting the encoder and decoder in each block. The encoder is made up of 3 residual blocks~\cite{DBLP:journals/corr/HeZRS15}, as depicted in Fig.~\ref{fig:arch_res_block}. These blocks each reduce the dimensions of the input by a factor of 2. The output of these layers is concatenated with the output of the corresponding \textit{upconv} layers in the decoder, which increase the scale of the input by a factor of 2. As depicted, these concatenated feature maps are then passed through \textit{conv} blocks. In this and subsequent diagrams, the number at the bottom of each cell indicates the number of feature maps output by this operation.

\begin{figure}
    \centering
    \includegraphics[width=0.35\linewidth]{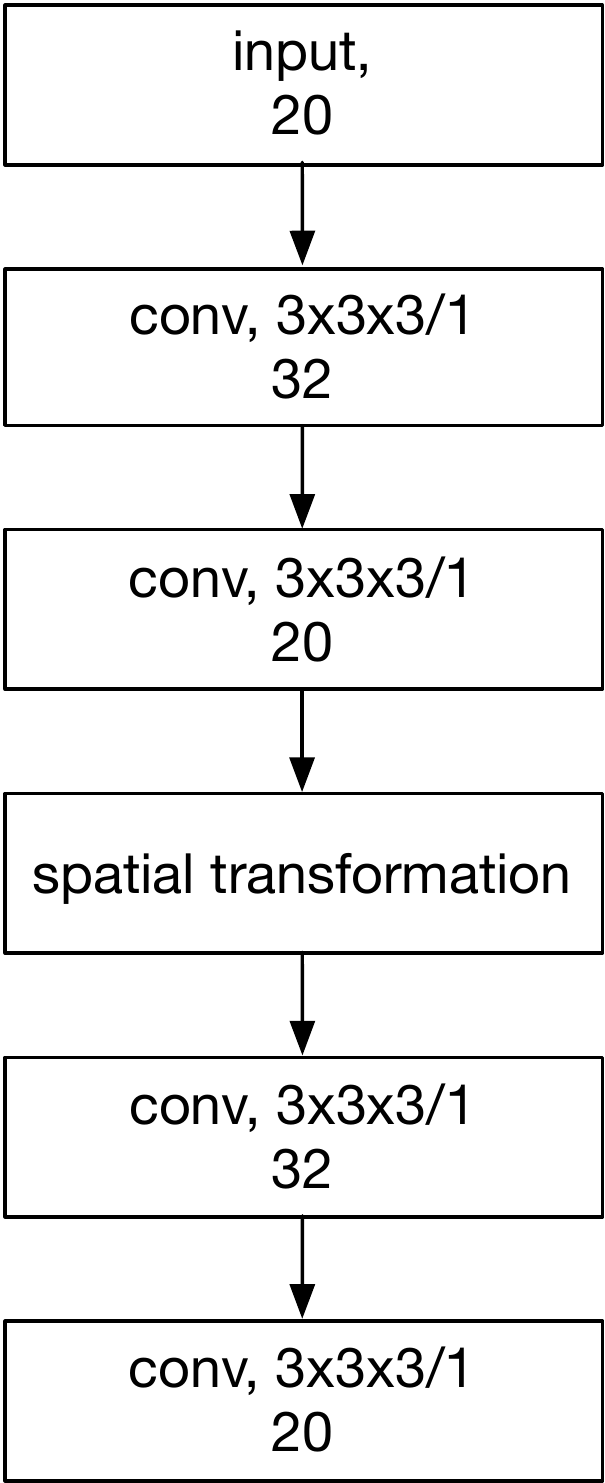}
    \caption{The 3D component of our network. The top row of the \textit{conv} blocks indicates the filter size and stride, while the number at the bottom of each block indicates the number of feature maps produced by the operation.}
    \label{fig:arch_3d_block}
\end{figure}

The architecture of the \textit{3d\_block} segment is depicted in Fig.~\ref{fig:arch_3d_block}. This block consists of 2 convolution layers ($3\times3$, stride $1$) applied before and after the spatial transformation.

\begin{table}[ht]
  \centering
  \resizebox{0.47\textwidth}{!}{
  \def\sym#1{\ifmmode^{#1}\else\(^{#1}\)\fi}
  \begin{tabular}{c*{6}{c}}
    \toprule
    & Layer Name & Output Size & Filter Size, Stride & Notes \\
    \midrule
    & input\_image & $160\times160\times3$ &  &  \\
    & conv & $80\times80\times32$ & $4\times4$, $2$ & \\
    & conv & $40\times40\times64$ & $7\times7$, $2$ & \\
    & unet\_block & $40\times40\times800$ &  & See Fig.~\ref{fig:arch_unet_block}\\
    & reshape & $40\times40\times40\times20$ & & Reshape 2D to 3D \\
    & 3d\_block & $40\times40\times40\times20$ &  & See Fig.~\ref{fig:arch_3d_block} \\
    & reshape & $40\times40\times800$ & & Reshape 3D to 2D \\
    & unet\_block & $40\times40\times20$ &  & See Fig.~\ref{fig:arch_unet_block} \\
    & upconv & $80\times80\times32$ &  & \\
    & conv & $80\times80\times32$ & $3\times3$, $1$ & \\
    & upconv & $160\times160\times32$ &  & \\
    & output & $160\times160\times4$ & $3\times3$, $1$ & \\
    \bottomrule
  \end{tabular}}
\caption{The architecture of our Transformable Bottleneck Network. Please consult the referenced figures for details on the individual segments of the network.}
\label{tab:arch}
\end{table}
 
\subsection{3D Reconstruction}
For the results provided for the 3D reconstruction task, we use the overall network structure described in Table~\ref{tab:arch}, except that we do not apply the first \textit{conv} and final \textit{upconv} layers, which halve and double the overall output dimensions, respectively. This results in a $32\times32\times32$ feature volume (with $20$ features per cell) when the network is applied to the $64\times64$ RGB images used as input to the network. This corresponds to the dimensions of the occupancy volume used in~\cite{tulsiani2017ray} and in our evaluations.

The network branch that serves as our occupancy decoder (see overview figure in the paper) has the same structure as the \textit{3d\_block} described above. However, in this case, the final 3D convolution layer produces only 1 feature per cell, and no further spatial transformation is applied in the middle of this block, as we are simply interested in obtaining the occupancy status for each cell in the feature volume. We apply a softmax operation in the depth dimension to the features produced by the occupancy decoder. In our experiments, we found that this softmax operation helped to normalize the input to a range that worked well for our reconstruction task, reducing the influence of extreme values in the occupancy volume.

To synthesize the 2D segmentation masks used for training, we reshape the occupancy volume into a $32\times32$ feature map with $32$ features per cell, then apply a $1\times1$ convolution with stride $1$ to these features to produce a single scalar feature per cell, followed by a sigmoid operation. This produces a 2D $32\times32$ segmentation mask with values between 0 and 1. This segmentation mask is then upsampled to the target resolution, $64\times64$. This mask is then used to compute the loss compared to the ground-truth segmentation masks from the dataset.

During training, this branch is applied to the feature volume immediately before the spatial transformation to obtain the occupancy volumes and segmentation masks corresponding to each source image, and after the feature volume aggregation and spatial transformation for the occupancy volume and segmentation mask corresponding to the target image.

For the 3D reconstruction evaluations, we generate target occupancy volumes aligned to the canonical view of the object used in the meshes that are voxelized to obtain the ground-truth occupancy volume for each object.

\section{3D Reconstruction Results}

In Table~\ref{tab:3dresults} we provide details on the results of the 3D reconstruction experiments described in the paper (Sec. 4.3, Fig. 5) and the comparison with those obtained by Tulsiani~\etal~\cite{tulsiani2017ray}.~\footnote{For a fair comparison, we report numbers obtained using the pre-trained models, datasets, and evaluation framework made available online by the authors for this work, which were overall somewhat lower than those reported in their paper.} We report the Intersection-over-Union (IoU, higher is better) between the reconstructed volume and the ground-truth results obtained by voxelizing the mesh rendered for the corresponding image. The top row provides the results obtained using our method and theirs for only one input image, from which we extract the corresponding occupancy volume. The subsequent rows present the results obtained using our method when using additional views and averaging the corresponding bottleneck layers (as is done when using multiple input images for novel view synthesis) before applying the occupancy decoder.

``real" indicates that additional views of the rendered object (chosen from the 10 renderings per object in the dataset used for evaluation) were used to create the occupancy volume. These results thus show how our method improves its results when the additional information provided by these views. ``synthetic" indicates that these additional views of the object under different poses were \emph{synthesized} by our encoder-decoder framework, given the single original image as input, before being passed through the encoder again and aggregated in the bottleneck with those from the other views. As such, the ``synthetic" results still rely on only a \emph{single} ``real" image as input. This allows for a fair comparison between our method and~\cite{tulsiani2017ray} in these cases.

``random poses" indicates that the azimuth and elevation for the synthesized viewpoints were selected at random from the same distributions as were used for rendering the training and evaluation sets. ``regular poses" indicates that these additional images were synthesized at regular intervals around the vertical axis. This allows the synthesized images to complement one another by providing contextual information that may be missing when poses are chosen at random. Our results demonstrate that using synthesized images with regular poses outperforms not only~\cite{tulsiani2017ray} and our method when using a single image, but even the use of real images at random poses. The reconstruction quality generally improves somewhat as additional views are synthesized, but using as little as 4 additional synthesized views, we obtain results that are superior to those obtained using each alternative we evaluated. This indicates that the generative power of our encoder-decoder framework can be used to create images that improve the overall quality of the structural information stored in the bottleneck produced by the encoder, when the encoded bottlenecks for these synthesized images are aggregated with that from the original input image.

We note that we obtain substantially better quantitative results on the chair and aero datasets, but obtain only slightly better results for the car dataset. We believe that this is due to the relatively simple and uniform structures of the objects in the car dataset, compared to the more varied shapes seen in the other datasets. The benefit obtained using our approach is more substantial for the latter datasets, in which simply producing a rough estimate of an average object's shape would result in larger errors than it would for the cars.

\begin{table}
  \centering
  \resizebox{0.47\textwidth}{!}{
  \def\sym#1{\ifmmode^{#1}\else\(^{#1}\)\fi}
  \begin{tabular}{c*{5}{c}}
    \toprule
    & Methods & \multicolumn{3}{c}{IoU}   \\
    \cmidrule(lr){3-5}
    & & \multicolumn{1}{c}{Chair} & \multicolumn{1}{c}{Car} & 
      \multicolumn{1}{c}{Aero} & \\
    \midrule
    \multirow{4}{*}{} & TBN & .3042 & .4664 & .2699       \\    
    & Tulsiani~\etal~\cite{tulsiani2017ray} &  .3913    &  .7113    &  .3332       \\
    \midrule
    \addlinespace
    \multirow{4}{*}{+1 view} & TBN, real, random poses & .3455 & .5233 & .3300       \\    
    & TBN, synthetic, random poses & .3387 & .5213 & .3251       \\
    & TBN, synthetic, regular poses & .3628 & .5727 & .3752        \\
    \midrule
    \addlinespace
    \multirow{4}{*}{\rotatebox[origin=c]{00}{+2 views}} & TBN, real, random poses & .3650 & .5479 & .3582       \\    
    & TBN, synthetic, random poses & .3532 & .5433 & .3474       \\
    & TBN, synthetic, regular poses & .3738 & .6025 & .4060        \\
    \midrule
    \addlinespace
    \multirow{4}{*}{+3 views} & TBN, real, random poses & .3753 & .5638 & .3741       \\    
    & TBN, synthetic, random poses & .3600 & .5573 & .3587       \\
    & TBN, synthetic, regular poses & .4312 & .6785 & .4490        \\
    \midrule
    \addlinespace
    \multirow{4}{*}{+4 views} & TBN, real, random poses & .3822 & .5754 & .3858       \\    
    & TBN, synthetic, random poses & .3648 & .5674 & .3668       \\
    & TBN, synthetic, regular poses & .4507 & .7128 & \textbf{.4661}        \\
    \midrule
    \addlinespace
    \multirow{4}{*}{+5 views} & TBN, real, random poses & .3878 & .5840 & .3941       \\    
    & TBN, synthetic, random poses & .3687 & .5748 & .3725       \\
    & TBN, synthetic, regular poses & .4455 & .7020 & .4498        \\
    \midrule
    \addlinespace
    \multirow{4}{*}{+6 views} & TBN, real, random poses & .3918 & .5913 & .4004       \\    
    & TBN, synthetic, random poses & .3714 & .5814 & .3768       \\
    & TBN, synthetic, regular poses & .4486 & .7075 & .4522        \\
    \midrule
    \addlinespace
    \multirow{4}{*}{+7 views} & TBN, real, random poses & .3946 & .5968 & .4049       \\    
    & TBN, synthetic, random poses & .3732 & .5862 & .3797       \\
    & TBN, synthetic, regular poses & .4546 & .7070 & .4530        \\
    \midrule
    \addlinespace
    \multirow{4}{*}{+8 views} & TBN, real, random poses & .3972 & .5996 & .4090       \\    
    & TBN, synthetic, random poses & .3748 & .5884 & .3827       \\
    & TBN, synthetic, regular poses & \textbf{.4630} & \textbf{.7131} & .4594        \\
    \midrule
    \addlinespace
    \multirow{4}{*}{+9 views} & TBN, real, random poses & .3988 & .6023 & .4132      \\    
    & TBN, synthetic, random poses & .3757 & .5906 & .3851       \\
    & TBN, synthetic, regular poses & .4561 & .7088 & .4565        \\
    \bottomrule
  \end{tabular}}
\caption{Quantitative results for 3D reconstruction using a single input image, and with up to 9 additional views (real or synthesized). We report the intersection-over-union (IoU, higher is better) for our method and Tulsiani~\etal~\cite{tulsiani2017ray}, which uses a single image as input.}
\label{tab:3dresults}
\end{table}
 
\section{Training}
The equation defining the total training loss is, as described in the paper,
\begin{align}
    \loss_{\mathrm{T}}(\allparams) = \loss_{\mathrm{R}} + \lambda_1 \loss_{\mathrm{P}}  + \lambda_2 \loss_\mathrm{S} + \lambda_3 \loss_\mathrm{A} + \lambda_4 \loss_\mathrm{M},
    \label{eq:total-loss}
\end{align}
where $\loss_{\mathrm{R}}$ is the $\L_1$ reconstruction loss, $\loss_{\mathrm{P}}$ is the $\L_2$ loss in the feature space of the VGG-19 network\footnote{We use the loss computed on the conv1\_1, conv2\_1, conv3\_1, and relu3\_3 layers of the VGG-19 network.}, $\loss_{\mathrm{S}}$ is the structural similarity (SSIM) index loss, $\loss_{\mathrm{A}}$ is the adversarial loss using the discriminator architecture from~\cite{tulyakov2018mocogan}), and $\loss_{\mathrm{M}}$ is the segmentation masking loss. Please see the paper for details on each of these loss terms. We empirically determined appropriate weights for the hyper-parameters controlling the contribution of the different loss components: $\lambda_1=5$, $\lambda_2=10$, $\lambda_3=0.05$, and $\lambda_4=10$.

We train the network using the Adam optimizer~\cite{DBLP:journals/corr/KingmaB14} with a learning rate set to $0.0002$, $\beta_1=0.9$ and $\beta_2=0.999$. Convergence on the test set typically takes approximately 8 days for each dataset we used for our evaluations.

\section{Datasets}
\subsection{Novel View Synthesis}
\subsubsection{ShapeNet Chairs and Cars}
We evaluate our framework's novel view synthesis (NVS) capabilities using the dataset provided for the benchmark in~\cite{sun2018nvs}.\footnote{The official code release, with pre-trained models and datasets, can be found at \href{https://github.com/shaohua0116/Multiview2Novelview}{https://github.com/shaohua0116/Multiview2Novelview}.} While the images were rendered at $256\times256$, our NVS network architecture accepts and produces images at a resolution of $160\times160$ for the $40\times40\times40$ volumetric bottleneck that we use for these evaluations\footnote{Using a larger volumetric bottleneck results in substantially higher memory usage and much longer training times}. We thus apply bilinear resampling to downsample the input and upsample the output to the resolution used during training. As this operation is differentiable, losses during training are measured with respect to the target image at its original resolution. We also report these losses used for the benchmark at the original target image resolution to make for a fair comparison to the other methods that we evaluated.

The car dataset consists of 5,997 models used for training and 1,500 used for testing. Rendering 54 views per each model~\footnote{18 azimuth angles sampled at 20-degree intervals and 3 elevations (0, 10 and 20 degrees).} results in 323,838 training images and 81,000 testing images. The chairs dataset consists of 558 training models and 140 testing models, resulting in 30,132 training images and 7,560 testing images.

Note that, while the training and testing images were rendered at 20-degree intervals around the vertical axis, in our supplementary video we provide examples of models rendered at 10-degree intervals. This demonstrates that our method is able to generalize to intermediate poses not seen during training. In contrast, for their ShapeNet evaluations, \cite{sun2018nvs} uses one-hot vectors indicating the discrete azimuth and elevation intervals at which the source images were rendered, and the specified pose for the target image. It is thus unclear how or whether their method would be able to generalize to intermediate poses not used for training.

Our NVS results for cars in the supplementary video also demonstrate that our network is able to synthesize transparent features such as the glass in the car windows.

\subsubsection{Human Action Dataset}
Each subject is rendered while performing 48 animation sequences, using rigged human models (varying in gender, ethnicity, size, age, and clothing) and animation sequences obtained from Renderpeople~\cite{Renderpeople:2018:RP}. For 4 frames selected at regular intervals in each animation sequence, the subjects are rendered at 12 viewpoints sampled at 30-degree intervals around the vertical axis. This results in 428,544 images. We use 128 subjects for training and the remaining 58 for evaluation, resulting in a total of 294,912 training images and 133,632 testing images.

While we use 30-degree increments for training on this dataset, in our supplementary video we provide synthesis results in which the subject is rendered at 15-degree intervals. This further demonstrates our method's generalization capabilities.

\subsection{3D Reconstruction}

To measure our framework's 3D reconstruction capabilities and compare it to recent work, we use the dataset and evaluation framework provided by~\cite{tulsiani2017ray}\footnote{The official code release, with pre-trained models and tools for generating these datasets and evaluating the reconstruction results, can be found at \href{https://github.com/shubhtuls/drc}{https://github.com/shubhtuls/drc}.}.

The dataset consists of rendered images of ShapeNet models from 3 object categories: chairs, cars and aeroplanes. We use 2831/810/404 models for training/testing/validation for the aeroplane dataset, 5247/1500/750 models for the car dataset and 4744/1356/678 models for the chair dataset. There are 10 images per each model, rendered with varying lighting conditions and the viewpoint azimuth and elevation uniformly sampled at random intervals in the ranges $[0,360)$ and $[-20,30]$, respectively.

While the images are rendered at a resolution of $224\times224$, we bilinearly downsample them to $64\times64$ for our network, which results in the $32\times32\times32$ occupancy volume that we use for evaluation. In contrast, we use images of size $160\times160$ and a $40\times40\times40$ feature volume for our novel view synthesis task. These 3D reconstruction results thus demonstrate that our network is able to extract meaningful structure from the input images even in the case of low input resolution and a smaller volumetric bottleneck resolution.

\section{Segmentation Supervision Ablation Study}
As discussed in the paper, we supervise our networks using a segmentation loss given the ground-truth foreground segmentation masks for each image. While this is useful for performing 3D reconstruction, to determine how crucial this supervision is for our approach to novel view synthesis we conducted an ablation study using a reduced version of our model. The architecture and training procedure is as described above, except that we use input images of a resolution of $128\times128$ and a bottleneck resolution of $32^3$. Random noise was used as the background for each input image. We found that our approach worked comparably well in reconstructing the foreground of the target evaluation images with and without this supervision.

Using the evaluation framework described in Sec. 4.2 for the chair dataset (using 4 input images for each target image), with segmentation supervision we achieved an average SSIM of 0.921 and an L1 loss (computed only for the foreground pixels of the target evaluation images) of 0.189. Without segmentation supervision, we achieved an SSIM of 0.920 and an L1 loss of 0.182. This suggests that, while useful for 3D reconstruction, this loss is not strictly necessary for novel view synthesis, as when it is omitted the network still learns to extract the features necessary to transform the foreground image content to the target view.

\section{Creative Manipulation Implementation.}

To perform spatial transformations to the encoded feature volume, we assign a $3$-dimensional coordinate $p_i=(x,y,z)$ to each cell $i$ corresponding to its spatial position in the volume, such that coordinate $(0,0,0)$ corresponds to the center of the bottleneck volume. During training, given an input and output image pair $\{ \image_k$, $\image_l \}$, we apply a rigid transformation corresponding to the relative pose of these images to these coordinates to determine the spatial position $p'_i$ in the transformed feature volume corresponding to $p_i$ in the original feature volume.
This provides us with the flow field $\flowfield_\ktol$ used to sample the encoded feature volume to produce the transformed feature volume that is passed to the decoder (see Sec. 3.1).

During training, we only apply rigid transformations corresponding to changes in the azimuth (rotations around the vertical axis, corresponding to the $y$-axis in our representation) and elevation (rotations around the horizontal $x$-axis) of the viewpoint of the scene.
However, in our results we demonstrate that this training process allows for performing \emph{non-rigid} transformations that enable a large variety of plausible manipulations to the image content.
Here we describe in more detail the method in which we perform these transformations to obtain the results seen in our paper (Figs. 1, 6-7) and the supplementary video.

\paragraph{Vertical and Horizontal Stretching.}
Rather than directly sampling from the region in the encoded volume based on the rigid pose between the input and output views as described, we can vary the sampling strategy to produce stretching effects such as those seen in our results (Fig. 1, row 4, Fig. 6, rows 1-2, and Fig. 7 in the main paper, and in the supplementary video, 2:50-3:40, 3:58-4:14, and 5:40-6:05).

For example, suppose that we have $n$ regularly sampled values corresponding to the $y$-positions of the cells in a slice of the transformed feature volume. We can find the corresponding values $(y_0,...,y_{n-1})$ to use when sampling the encoded feature volume as follows:

\begin{equation}
y_i= a + \frac{b-a}{n-1}\times{i}
\end{equation}


Thus we have $y_0=a$ and $y_{n-1}=b$. By adjusting $a$ and $b$ we can change the position and size of the region in the input volume from which values are sampled, which will alter how the transformed region is compressed or stretched in the decoded image. We use this technique with multiple slices with input positions that vary over time to produce the vertically stretching chair animations in our results. (After sampling the input volume as described, we apply a rigid rotation to produce the novel viewpoints seen in these images.)


We can apply similar techniques to produce stretching effects in the $x$- and $z$-dimensions. By varying the slice parameters over time, we cause the cars seen in our supplementary video to vary in length and width over the course of the animation.

\paragraph{Vertical Twisting.}
We can also apply different rigid rotations to separate regions of the encoded feature volume, we can achieve the ``twisting'' effect seen in our results (Fig. 6, row 3 in the main paper and in the supplementary video, 2:50-3:40). Given a user-specified point on the vertical $y$-axis and a rotation value $\alpha$, we apply a rigid rotation around the $y$-axis to the feature volume of $\alpha$ degrees for all cells above this point and $-\alpha$ degrees for all cells below this point. Varying the $\alpha$ parameter over time produces the twisting effect seen in the swivel chairs portrayed in our results.

\paragraph{Volume Merging and Reflection.}
By combining the content of different regions of multiple encoded feature volumes, we can decode images in which this content has been merged in a corresponding fashion as seen in Fig. 1 of the main paper (row 4, columns 4-6) and in the supplementary video (4:14-4:40). In these examples, the top and bottom halves of the feature volumes for 2 different individuals have been combined to produce new subjects with an appearance corresponding to the upper half of the first subject and the lower half of the second. Note that while this appears to produce an effect similar to that of merging the upper and lower regions of the rendered images, after performing this merging \emph{once} for the encoded bottlenecks for each subject, we can rigidly transform the result to produce novel views of the subject, as seen in the supplementary video.

Similarly, we can alter and replicate regions of a single bottleneck to produce novel content, as seen in the slicing and stitching examples in the main paper (Fig. 6, rows 4-5) and in the supplementary video (3:40-3:58). For these examples, we discard the feature volume content contained within one half of the $xy$-plane, and reflect the content of the cells in the remaining half across this plane to fill the missing regions. As a result we can produce new shapes in which the front of the depicted car has been replaced by the back (Fig. 6, row 4), or vice-versa (Fig. 6, row 5). As before, we can also apply rigid transformations to the result to produce novel views of the image content. Interestingly, these rendered results still plausibly produce the specified manipulation of the encoded feature volume, though cars with such unusual shapes were never seen during training.

\end{document}